\documentclass[a4paper,conference]{IEEEtran}
\usepackage{times}
\usepackage[utf8]{inputenc} 
\usepackage[T1]{fontenc}    
\usepackage{url}            
\usepackage{booktabs}       
\usepackage{amsfonts}       
\usepackage{nicefrac}       
\usepackage{microtype}      
\usepackage{lipsum}
\usepackage{xcolor}
\usepackage{subfig}
\usepackage{graphicx}       
\usepackage{caption}
\usepackage{epsfig}
\usepackage{amsmath}
\usepackage{amssymb}
\usepackage{float}
\usepackage{tabularx}
\usepackage{multirow}

\usepackage{mathtools, nccmath}
\DeclarePairedDelimiter{\nint}\lfloor\rceil

\newcommand{\etal}{et al.}
\definecolor{lightergray}{rgb}{0.8, 0.8, 0.8}

\begin{document}


%

\title{Knowledge Distillation Beyond Model Compression}

\author{%
  \IEEEauthorblockN{%
    Fahad Sarfraz\IEEEauthorrefmark{1},
    Elahe Arani\IEEEauthorrefmark{1} and
    Bahram Zonooz
  }%
  \IEEEauthorblockA{Advanced Research Lab, NavInfo Europe, Eindhoven, The Netherlands}%
}

\maketitle
\begingroup\renewcommand\thefootnote{\IEEEauthorrefmark{1}}
\footnotetext{Equal contribution}
\endgroup

\maketitle

\begin{abstract}
Knowledge distillation (KD) is commonly deemed as an effective model compression technique in which a compact model (student) is trained under the supervision of a larger pretrained model or an ensemble of models (teacher).
Various techniques have been proposed since the original formulation, which mimic different aspects of the teacher such as the representation space, decision boundary, or intra-data relationship. Some methods replace the one-way knowledge distillation from a static teacher with collaborative learning between a cohort of students.
Despite the recent advances, a clear understanding of where knowledge resides in a deep neural network and an optimal method for capturing knowledge from teacher and transferring it to student remains an open question.
In this study, we provide an extensive study on nine different KD methods which covers a broad spectrum of approaches to capture and transfer knowledge. We demonstrate the versatility of the KD framework on different datasets and network architectures under varying capacity gaps between the teacher and student. The study provides intuition for the effects of mimicking different aspects of the teacher and derives insights from the performance of the different distillation approaches to guide the design of more effective KD methods.  
Furthermore, our study shows the effectiveness of the KD framework in learning efficiently under varying severity levels of label noise and class imbalance, consistently providing generalization gains over standard training. We emphasize that the efficacy of KD goes much beyond a model compression technique and it should be considered as a general-purpose training paradigm which offers more robustness to common challenges in the real-world datasets compared to the standard training procedure.
\end{abstract}

\section{INTRODUCTION}
Deep convolutional neural networks (DNNs) have achieved state of the art performance in many visual recognition tasks \cite{krizhevsky2012imagenet, voulodimos2018deep}.
However, the state of the art performance comes at the cost of training computationally expensive and memory-intensive networks with large depth and/or width with tens of millions of parameters and requires billions of operations per inference.
This hinders their deployment in resource-constrained devices or in applications with strict latency requirements such as self driving cars and hence leads to a necessity for developing compact networks that generalize well.

Several model compression techniques have been proposed such as model quantization \cite{zhou2017stochastic}, model pruning \cite{han2015deep}, and knowledge distillation \cite{hinton2015distilling}, each of which has its own set of advantages and drawbacks \cite{cheng2017survey}.
Amongst these techniques, our study focuses on knowledge distillation (KD) as it involves training a smaller compact network (student) under the supervision of a larger pre-trained network or an ensemble of models (teacher) in an interactive manner which is more similar to how humans learn.
KD has proven to be an effective technique for training a compact model and also provides greater architectural flexibility since it allows structural differences in the teacher and student.
Furthermore, since it provides a training paradigm instead of a method for compressing a trained model, we aim to analyze its characteristics as a general-purpose training framework beyond just model compression.

In the original formulation, Hinton \etal~\cite{hinton2015distilling} proposed mimicking the softened softmax output of the teacher.
Since then, a number of KD methods have been proposed, each trying to capture and transfer some characteristic of the teacher such as the representation space, decision boundary or intra-data relationship.
However, a clear understanding of where knowledge resides in a deep neural network is still lacking and consequently an optimal method of capturing knowledge from teacher and transferring it to student remains an open question.
Furthermore, it is plausible to argue that the effectiveness of the distillation approach is dependent upon a number of factors: the capacity gap between the student and teacher, the nature and degree of the constraint put on student training, and the characteristic of the teacher mimicked by the student. 
Hence, it is important to extensively study the effectiveness and versatility of different KD methods capturing different aspects of the teacher under a uniform experimental setting to gain further insights.

Furthermore, the standard training procedure with the cross-entropy loss on one-hot-encoded labels has a number of shortcomings.
Over-parameterized DNNs trained with standard cross-entropy loss have been shown to have the capacity to fit any random labeling of data which makes it challenging to learn efficiently under label noise \cite{hu2019understanding, arpit2017closer}.
Also, when class imbalance exists within the training data, DNNs tend to over-classify the majority group due to its increased prior probability \cite{johnson2019survey}.
We hypothesize that one key factor contributing to these inefficiencies is that the only information the model receives about the classes is the one-hot-encoded labels and therefore it fails to capture additional information about the structural similarities between different classes.
The additional supervision in KD about the relative probabilities of secondary classes and/or relational information between data points can be useful in increasing the efficacy of the network to learn under label noise and class imbalance.
To test the hypothesis we simulate varying degrees of label noise and class imbalance and show how different KD methods perform under these challenging situations.

Our empirical results demonstrate that the effectiveness of the KD framework goes beyond just model compression and should be further explored as an effective general-purpose learning paradigm.

Our main contributions are as follows:
\begin{enumerate}
    \item A study of nine different KD methods on different datasets (CIFAR-10 and CIFAR-100) and network architectures (ResNets and WideResNets) with varying degree of capacity gaps between the teacher and student.
    \item Demonstrate KD as an effective approach for learning under label noise and class imbalance.
    \item Provide insights into the effectiveness of mimicking different aspects of teacher.
\end{enumerate}

\section{Knowledge Distillation Methods}
KD aims to improve the performance of the student by providing additional supervision from the teacher. 
A number of methods have been proposed to decrease the performance gap between student and teacher.
However, what constitutes knowledge and how to optimally transfer it from the teacher is still an open question.
Here, we cover a diverse set of KD methods which differ from each other with respect to how knowledge is defined and transferred from the teacher. 
To highlight the subtle differences among the distillation methods used in the study, we present a broad categorization of these methods.

\paragraph{Response Distillation}
aims to mimic the output of the teacher.
The key idea is that student can be trained to generalize the same way as the teacher by using the output probabilities produced by the teacher as a "soft target".
The relative probabilities of the incorrect labels encode a rich similarity structure over the data and holds information about how the teacher generalizes.
Bucilua \etal~\cite{bucilua2006model} originally proposed to use the logits instead of the probabilities as target for the student and to minimize the squared difference.
Hinton \etal~\cite{hinton2015distilling} built upon the previous work and proposed to raise the temperature of the final softmax function and minimize the Kullback–Leibler (KL) divergence between the smoother output probabilities.
BSS~\cite{heo2019knowledge} proposed a method for more explicitly matching the decision boundary by utilizing an adversarial attack to discover samples supporting a decision boundary and use an additional boundary supporting loss which encourages the student to match the output of the teacher on samples close to the decision boundary.
Response distillation can be seen as an implicit method for matching the decision boundaries of the student and the teacher.

\paragraph{Representation Space Distillation} 
aims to mimic the latent feature space of the teacher.
FitNet~\cite{romero2014fitnets} introduced intermediate-level hints from the teacher's hidden layers to guide the training process of the student.
This encourages the student to learn an intermediate representation that is predictive of the intermediate representations of the teacher network.
Instead of mimicking the intermediate layer activation values, which can be viewed as putting a hard constraint on the student training, FSP~\cite{yim2017gift} proposed to capture the transformation of features between the layers. 
The method encourages the student to mimic the teacher’s flow matrices, which are derived from the inner product between feature maps in two layers. 
Both FitNet and FSP employ a two-stage training procedure whereby the first stage involves mimicking the representation of the teacher for better initialization of the student and the second stage involves training the entire model with Hinton's method.
AT~\cite{zagoruyko2016paying} proposed attention as a mechanism of transferring knowledge. They define attention as a set of spatial maps that encode on which spatial areas of the input, the network focuses most for taking its output decision based on the activation values.
Compared to FitNet, AT and FSP can be considered as softer constraints on the student training.
AT, unlike FitNet and FSP, puts a constraint on the student to mimic the representation space (i.e., attention maps) throughout the training procedure and is therefore perhaps more representative of the representation space distillation.  

\paragraph{Relational Knowledge Distillation}
aims to mimic the structural relations between the learned representation of the teacher using the mutual relations of data samples in the teacher's output representation. 
RKD~\cite{park2019relational} emphasizes that knowledge within a neural network is better captured by the relations of the learned representation than the individuals of those. 
Their distillation approach trains the student to form the same relational structure with that of the teacher in terms of two variants of relational potential functions:
Distance-wise potential, RKD-D, measure the Euclidean distance between two data samples in the output representation space and angle wise potential, RKD-A, measures the angle formed by the three data samples in the output representation space. RKD-DA combines both of these losses to train the student.
SP~\cite{tung2019similarity} encourages the student to preserve the pairwise similarities in the teacher in such a way that data pairs that produce similar/dissimilar activations in the teacher, also produce similar/dissimilar activations in the student. 
SP can be viewed as an instance of RKD angle wise potential variant and differs from RKD-A in that, it uses the dot product i.e. cosine angle between pairs of data.
Relational knowledge distillation does not require the student to mimic the representation space of the teacher and hence provides more flexibility to the student.

\paragraph{Online Knowledge Distillation} 
aims to circumvent the need for a static teacher and the associated computational cost.
Deep Mutual Learning (DML)~\cite{zhang2018deep} replaces the one way knowledge transfer from a pretrained model with knowledge sharing between a cohort of compact models trained collaboratively.
DML involves training each student with two losses: a conventional supervised learning loss, and a mimicry loss that aligns
each student’s class posterior with the class probabilities of other students.
To address the limitation of lacking a high capacity model and the need for training multiple student networks, ONE~\cite{lan2018knowledge} uses a single multi-branch network and uses an ensemble of the branches as a stronger teacher to assist the learning of the target network.


\begin{table}
\centering
\caption{\footnotesize Training parameters for each of the KD method. Distillation layer refers to the position in the network where the distillation loss is applied. Block n means the output of the n-th ResNet/WRN block and the output layer refers to the final softmax output.}
\label{tab:train-params}
\begin{tabular}{p{0.04\textwidth}p{0.13\textwidth}p{0.08\textwidth}p{0.13\textwidth}} \hline
& \textbf{Training Scheme} & \textbf{Distillation Layer} & \textbf{Parameters}\\\hline\hline
\bf Hinton & Standard & Output Layer & $\alpha=0.9$, $T=4$\\ \hline
\bf BSS & 200 epochs & Output Layer & \begin{tabular}[c]{@{}l@{}}T=4\\ attack size = 32\\ num steps =10\\ max eps = 16\end{tabular}\\ \hline
\bf FitNet & \begin{tabular}[c]{@{}l@{}}Stage 1: Initialization\\ Stage 2: Standard\end{tabular} & Block 2 & \begin{tabular}[c]{@{}l@{}}Stage 1: W = {[}0, 1, 0{]}\\ Stage 2: $\alpha=0.9$,\\ ~~~~~~~~~~~$T=4$\end{tabular} \\ \hline
\bf FSP & \begin{tabular}[c]{@{}l@{}}Stage 1: Initialization\\ Stage 2: Standard\end{tabular} & Blocks 1,2,3 & \begin{tabular}[c]{@{}l@{}}Stage 1: W = {[}1, 3, 1{]}\\ Stage 2: $\alpha=0.9$,\\ ~~~~~~~~~~~$T=4$\end{tabular} \\ \hline
\bf AT & Standard & Blocks 1,2,3 & $\beta=1000$ \\ \hline
\bf SP & Standard & Block 3 & $\gamma=3000$ \\ \hline
\bf \bf \bf RKD & Standard & Block 3 & \begin{tabular}[c]{@{}l@{}}$\lambda_{RKD-D}=25$\\ $\lambda_{RKD-A}=50$ \end{tabular}\\ \hline
\bf \bf ONE & As per paper & N/A & T=3\\ \hline
\bf DML & Standard & Output Layer & T=4, Cohort size = 2 \\ \hline
\end{tabular}
\end{table}

\section{Experimental Setup}
For our empirical analysis, we perform our experiments on CIFAR-10 and CIFAR-100 datasets \cite{krizhevsky2010cifar} with ResNet \cite{he2015deep} and Wide Residual Networks (WRN) \cite{zagoruyko2016wide} and evaluate the efficiency of the different distillation methods on varying capacity gaps between the teacher and student. 
To have a fair comparison between the different KD methods, we use a consistent set of hyperparameters and training scheme where possible. 
Unless otherwise stated, for all our experiments we use the following training scheme as used in Zagoruyko \etal~\cite{zagoruyko2016paying}:
normalize the images between 0 and 1; random horizontal flip and random crop data augmentations with reflective padding of 4; Stochastic Gradient Descent with 0.9 momentum; 200 epochs; batch size 128; and an initial learning rate of 0.1, decayed by a factor of 0.2 at epochs 60, 120, and 150. 
For the network initialization stage required for FitNet and FSP, we use 100 epochs, and an initial learning rate of 0.001, decayed by a factor of 0.1 at epochs 60 and 80. Table \ref{tab:train-params} provides the details of the training parameters for each distillation method.
We train all the models for 5 different seed values.
For the teacher, we select the model with the highest test accuracy and then use it to train the student for five different seed values and report the mean value for our evaluation metrics. We use the classification branch, and student model with the highest accuracy on the test dataset for ONE and DML methods respectively for each seed.
For the online distillation methods, ONE has the student architecture with an ensemble of three classification branches providing teacher supervision and DML uses mutual learning between two models with the same student architecture.
For all the other KD methods, we use WRN-40-2 and ResNet-26 as the teacher for the WRN and ResNet student models respectively.
Note, that for the methods which report the results for the same student and teacher network configuration \cite{zagoruyko2016paying, heo2019knowledge, tung2019similarity}, our rerun under the aforementioned experimental setup achieves superior results than reported in the original work.

\begin{table}
\centering
\caption{\footnotesize Teacher performance for the different experiments. Default configuration is used for the experiments in table \ref{tab:cifar10-test-acc} and \ref{tab:cifar100-test-acc}. Rows 5-7 provides teacher for different label noise rates, $\sigma$, used in table \ref{tab:label_noise}, whereas rows 8-11 provides the teacher performance for different imbalance rates, $\gamma$, used in table \ref{tab:class-imbalance}.} 
\label{tab:teachers}
\begin{tabular}{lll|c}
\hline
\bf Dataset & \textbf{Architecture} & \textbf{Configuration} & \textbf{Accuracy (\%)}  \\ \hline \hline
\multirow{2}{*}{CIFAR-10}   & \multirow{1}{*}{ResNet-26}  & default           & 93.83\\
                            & \multirow{1}{*}{WRN-40-2}   & default           & 95.11\\\hline
\multirow{2}{*}{CIFAR-100}  & \multirow{1}{*}{ResNet-26}  & default           & 79.22\\ 
                            & \multirow{1}{*}{WRN-40-2}   & default           & 76.06 \\ \hline
\multirow{7}{*}{CIFAR-10}   & \multirow{3}{*}{WRN-40-2}   & $\sigma=$ 0.2     & 82.66\\
                            &                             & $\sigma=$ 0.4     & 66.34\\
                            &                             & $\sigma=$ 0.6     & 45.97\\ \cline{2-4}
                            & \multirow{4}{*}{WRN-40-2}   & $\gamma=$ 0.2     & 78.87\\
                            &                             & $\gamma=$ 0.6     & 80.23\\
                            &                             & $\gamma=$ 1.0     & 81.13\\
                            &                             & $\gamma=$ 2.0     & 84.40\\ \hline
\end{tabular}
\end{table}


\section{Empirical Analysis}
The aim of the study is manifold:
a) provide extensive analysis of how the underlying mechanisms of different KD methods affect the generalization performance of the student under uniform experimental conditions.
b) Demonstrate the versatility of the KD framework on different datasets and network architectures under varying capacity gaps between the teacher and student.
c) Highlight the efficacy of KD framework as a general-purpose training framework which provides additional benefits over model compression.
To this end, section \ref{gen_perf} compares the performance of the KD methods across different datasets and architectures. 
Section \ref{label_noise} evaluates the efficiency of KD methods under various degrees of label noise. 
Section \ref{class_imbalance} further evaluates the performance on imbalanced datasets. 
Finally, section \ref{transferability} studies the transferability of adversarial examples between the student models trained with different KD methods.

It is important to note that the goal of our study is not to rank different KD methods based on the test set performance, but instead to provide a comprehensive evaluation of the methods and find general patterns and characteristics of the model to aid the design of efficient distillation methods. In addition, we aim to showcase the efficacy of the KD framework as a robust general-purpose learning paradigm. 

\begin{table*}[tb]
\centering
\caption{\footnotesize Test set performance (\%) on CIFAR-10. The best results are in bold. We run each experiment for 5 different seeds and report the mean $\pm$ 1 STD.}
\label{tab:cifar10-test-acc}
\resizebox{\textwidth}{!}{%
\begin{tabular}{|l|cccc||cccc|}
\hline
 & \textbf{ResNet-8} & \textbf{ResNet-14} & \textbf{ResNet-20} & \textbf{ResNet-26} & \textbf{WRN-10-2} & \textbf{WRN-16-2} & \textbf{WRN-28-2} & \textbf{WRN-40-2} \\ \hline \hline
\textbf{Baseline} & 87.64$\pm$0.25 & 91.44$\pm$0.15  & 92.64$\pm$0.18  & 93.32$\pm$0.37  & 90.62$\pm$0.15 & 93.95$\pm$0.18 & 94.82$\pm$0.10 & 95.01$\pm$0.11 \\ \hline
\textbf{Hinton}   & 88.80$\pm$0.16 & \bf 92.50$\pm$0.19  & 93.25$\pm$0.18  & 93.58$\pm$0.10  & 91.72$\pm$0.12 & 94.28$\pm$0.09 & 94.97$\pm$0.10 & 95.12$\pm$0.10 \\ 
\textbf{BSS}      & 89.18$\pm$0.43 & 91.99$\pm$0.20  & 92.92$\pm$0.18  & 93.52$\pm$0.08  & \bf 92.32$\pm$0.21 & 94.27$\pm$0.18 & 94.72$\pm$0.15 & 94.96$\pm$0.20 \\ \hline
\textbf{FitNet}   & 88.89$\pm$0.21 & \bf 92.50$\pm$0.10  & 93.27$\pm$0.15  & 93.58$\pm$0.10  & 91.65$\pm$0.08 & 94.34$\pm$0.11 & 94.94$\pm$0.14 & 95.10$\pm$0.14 \\
\textbf{FSP}      & 88.77$\pm$0.41 & 92.18$\pm$0.19  & 93.29$\pm$0.30  & 93.73$\pm$0.16  & 91.70$\pm$0.26 & 94.31$\pm$0.08 & 95.06$\pm$0.19 & 95.15$\pm$0.19 \\ 
\textbf{AT}       & 86.07$\pm$0.32 & 91.66$\pm$0.16  & 92.96$\pm$0.09  & 93.32$\pm$0.14  & 90.99$\pm$0.21 & 94.50$\pm$0.18 & \bf 95.32$\pm$0.20 & 95.39$\pm$0.15 \\ \hline
\textbf{SP}       & 86.62$\pm$0.26 & 92.34$\pm$0.19  & 93.28$\pm$0.07  & 93.70$\pm$0.23  & 91.27$\pm$0.26 & \bf 94.64$\pm$0.17 & 95.25$\pm$0.14 & 95.35$\pm$0.11 \\ 
\textbf{RKD-D}    & 87.48$\pm$0.21 & 91.87$\pm$0.19  & 92.94$\pm$0.30  & 93.56$\pm$0.16  & 90.99$\pm$0.17 & 94.42$\pm$0.15 & 95.09$\pm$0.08 & 95.31$\pm$0.13 \\ 
\textbf{RKD-A}    & 87.32$\pm$0.24 & 92.01$\pm$0.14  & \bf 93.30$\pm$0.12  & 93.67$\pm$0.13  & 90.98$\pm$0.31 & 94.62$\pm$0.14 & 95.23$\pm$0.13 & 95.36$\pm$0.27 \\ 
\textbf{RKD-DA}   & 87.14$\pm$0.19 & 92.05$\pm$0.20  & 93.05$\pm$0.20  & 93.73$\pm$0.09  & 90.92$\pm$0.16 & 94.52$\pm$0.11 & 95.19$\pm$0.12 & \bf 95.41$\pm$0.07 \\ \hline
\textbf{ONE}      & \bf 89.54$\pm$0.17 & 92.30$\pm$0.23  & 93.27$\pm$0.16  & \bf 93.80$\pm$0.13  & 87.75$\pm$1.92 & 92.80$\pm$0.08 & 94.70$\pm$0.18 & 95.11$\pm$0.09 \\ 
\textbf{DML} & 87.94$\pm$0.15 & 92.20$\pm$0.18  & 93.14$\pm$0.06  & 93.45$\pm$0.10  & 91.60$\pm$0.28 & 94.38$\pm$0.15 & 95.17$\pm$0.10 & 95.33$\pm$0.09 \\ \hline
\end{tabular}%
}
\end{table*}

\subsection{Generalization Performance}\label{gen_perf}
KD aims to minimize the generalization gap between the teacher and the student.
Therefore, the generalization gain over the baseline (a model trained without teacher supervision) is a key metric for evaluating the effectiveness of a KD method.
Tables \ref{tab:cifar10-test-acc} and \ref{tab:cifar100-test-acc} demonstrate the effectiveness and versatility of the different KD methods in improving the generalization performance of the student on CIFAR-10 and CIFAR-100 datasets, respectively. 
For the majority of the methods, we see generalization gain over the baseline.

\subsubsection{\bf Results on CIFAR-10}

Amongst the \textbf{response distillation methods}, Hinton consistently improves the performance of the student over the baseline and proves effective across varying degrees of capacity gaps. 
BSS is effective for lower capacity students, providing the highest generalization for WRN-10-2, but it adversely affects the performance as the capacity of the model increases.

For the \textbf{representation space distillation} methods, FitNet and FSP provide similar performance to Hinton owing to the two-stage training scheme whereby the second stage is essentially the Hinton method. Both of these methods improve the generalization over the baseline consistently, however occasionally failing to outperform the Hinton method. AT, fails to improve over Hinton for the ResNet models, even decreasing the performance over the baseline for ResNet-8. For higher capacity WRN models, however, AT proves to be much more effective, and is amongst top-performing techniques for WRN-16-2 and WRN-28-2.

\textbf{Relational knowledge distillation} methods also show promising results, with SP and RKD-A amongst the top performing models for ResNet-20, WRN-16-2 and higher capacity variants. RKD-DA provides the highest generalization for WRN-40-2 and close to the highest value for ResNet-26. However, the effectiveness of these methods drops substantially as the capacity gap between the teacher and student increases e.g. for ResNet-8, all the relational distillation methods decreases the performance over the baseline.  

Lastly, \textbf{online distillation} demonstrates comparable effectiveness in improving the performance of the baseline model to their traditional counterparts which uses a static teacher model for supervision. ONE consistently improves the generalization of the baseline for ResNet models and provides the highest performance for ResNet-8 and ResNet-26. However, for WRN, it only manages to improve the performance for the higher capacity WRN-40-2 model. DML consistently provides generalization gains over baseline across all the models.
The results of ONE (on ResNet) and DML are promising given that they do not have the advantage of supervision from a high performing teacher model as for the other distillation methods.

\subsubsection{\bf Results on CIFAR-100}
The performance of the KD methods on CIFAR-100 follows a similar pattern observed on CIFAR-10 with a few deviations. 
Response distillation methods are among the most efficient methods across the different models.
Hinton improves the generalization for all the models except WRN-10-2.
BSS again proves to be effective on lower capacity models, providing the highest generalization for WRN-10-2.
Similar to CIFAR-10, FitNet and FSP provide similar performance to Hinton, and AT is detrimental for lower capacity models.
Relational knowledge distillation methods exhibit similar behavior as for CIFAR-10, proving more effective when the capacity gap is less. SP provides the highest generalization for ResNet-14 and higher capacity ResNet models.
Online distillation methods also exhibit similar behavior, with ONE proving to be effective on ResNet but failing to improve the performance on WRN except for WRN-40-2.
DML, on the other hand, is particularly effective on WRN, providing the highest generalization gains on WRN-16-2 and higher capacity WRN models.
\begin{table*}[tb]
\centering
\caption{\footnotesize Test set performance (\%) on CIFAR-100. The best results are in bold. We run each experiment for 5 different seeds and report the mean $\pm$ 1 STD.}
\label{tab:cifar100-test-acc}
\resizebox{\textwidth}{!}{%
\begin{tabular}{|l|cccc||cccc|}
\hline
 & \textbf{ResNet-8} & \textbf{ResNet-14} & \textbf{ResNet-20} & \textbf{ResNet-26} & \textbf{WRN-10-2} & \textbf{WRN-16-2} & \textbf{WRN-28-2} & \textbf{WRN-40-2} \\ \hline\hline
\textbf{Baseline} & 71.78$\pm$0.26 & 76.95$\pm$0.43  & 77.92$\pm$0.40  & 78.82$\pm$0.24  & 67.99$\pm$0.55 & 72.35$\pm$0.36 & 74.93$\pm$0.39 & 75.94$\pm$0.12 \\\hline
\textbf{Hinton}   & 72.78$\pm$0.36 & 78.18$\pm$0.14  & 79.58$\pm$0.22  & 79.73$\pm$0.30  & 67.70$\pm$0.54 & 74.12$\pm$0.37 & 76.08$\pm$0.36 & 77.07$\pm$0.13 \\
\textbf{BSS}      & 73.02$\pm$0.07 & 76.96$\pm$0.19  & 78.37$\pm$0.27  & 78.66$\pm$0.23  & \bf 69.55$\pm$0.38 & 73.04$\pm$0.21 & 75.59$\pm$0.24 & 76.55$\pm$0.16 \\ \hline
\textbf{FitNet}   & 72.86$\pm$0.28 & 78.48$\pm$0.30  & 79.55$\pm$0.18  & 79.83$\pm$0.34  & 67.83$\pm$0.47 & 72.82$\pm$2.40 & 76.31$\pm$0.09 & 77.25$\pm$0.15 \\
\textbf{FSP}      & 72.93$\pm$0.24 & 78.34$\pm$0.45  & 79.65$\pm$0.18  & 79.62$\pm$0.18  & 67.64$\pm$0.19 & 73.86$\pm$0.27 & 76.21$\pm$0.14 & 77.09$\pm$0.27 \\
\textbf{AT}       & 71.99$\pm$0.08 & 76.88$\pm$0.20  & 78.35$\pm$0.10  & 78.94$\pm$0.34  & 67.45$\pm$0.27 & 72.78$\pm$0.32 & 75.51$\pm$0.13 & 76.60$\pm$0.13 \\\hline
\textbf{SP}       & 73.18$\pm$0.24 & \bf 78.53$\pm$0.29  & \bf 79.76$\pm$0.27  & \bf 79.93$\pm$0.29  & 66.77$\pm$0.27 & 73.42$\pm$0.37 & 76.52$\pm$0.35 &  77.43$\pm$0.14 \\
\textbf{RKD-D}    & 71.99$\pm$0.23 & 77.02$\pm$0.21  & 78.23$\pm$0.26  & 78.80$\pm$0.28  & 68.14$\pm$0.34 & 72.52$\pm$0.30 & 75.48$\pm$0.33 & 76.34$\pm$0.29 \\
\textbf{RKD-A}    & 71.95$\pm$0.33 & 76.93$\pm$0.35  & 78.51$\pm$0.25  & 79.10$\pm$0.18  & 68.10$\pm$0.31 & 72.87$\pm$0.23 & 75.50$\pm$0.41 & 76.97$\pm$0.17 \\
\textbf{RKD-DA}   & 71.70$\pm$0.19 & 77.14$\pm$0.40  & 78.64$\pm$0.21  & 79.16$\pm$0.11  & 67.94$\pm$0.37 & 72.88$\pm$0.23 & 75.73$\pm$0.32 & 76.91$\pm$0.22 \\\hline
\textbf{ONE}      & 73.30$\pm$0.12 & 78.04$\pm$0.07  & 79.24$\pm$0.18  & 79.74$\pm$0.27  & 57.38$\pm$2.11 & 69.78$\pm$0.94 & 74.49$\pm$0.54 & 76.89$\pm$0.27 \\
\textbf{DML} & \bf 73.57$\pm$0.09 & 78.07$\pm$0.20  & 79.15$\pm$0.22 & 79.32$\pm$0.38 & 68.99$\pm$0.23 & \bf 74.44$\pm$0.25 & \bf 76.65$\pm$0.17 & \bf 77.65$\pm$0.19 \\ \hline
\end{tabular}%
}
\end{table*}

\subsubsection{\bf Key Insights}
From the empirical study, we derive the following insights, which can provide some guidelines for designing effective KD methods.
\begin{enumerate}
    \item KD framework is an effective technique which consistently provides generalization gains. The methods are generally effective and versatile enough to cater to different datasets and network architectures even for the higher capacity gap between the student and teacher. KD is not only effective for model compression but also improves the performance of the model when the student and the teacher have the same network architecture.
    \item The original Hinton method, while simple, is quite effective and versatile, providing comparable performance gains to the recently proposed distillation methods. This shows the utility of response distillation. It also provides more architectural flexibility between the student and the teacher.
    \item The performance of relational knowledge distillation methods provides a compelling case for the effectiveness of using the relations of the learned representations for KD. The comparison between SP and RKD-D is particularly interesting since both methods use pair-wise similarities at the same network layers. SP captures the angular relationship between the two vectors whereas RKD-D uses euclidean distance. 
    RKD-A measures the angle formed between triplets of data points and provides similar performance to SP. 
    The results suggest that angular information can capture higher-level structure which aids in a performance gain.
    \item Considering AT as a better representative for representation space distillation, the constraints put by these methods on the learned representation can be detrimental to the performance of the student when the capacity gap is high. For a substantially lower capacity model, mimicking the representation space of the teacher with much higher capacity might be difficult and perhaps not the optimal approach. Generally, we observe that the methods which provide more flexibility to the student in learning e.g. response distillation and relational KD methods are more versatile and can provide higher performance gains.
    \item The occasional performance gains with FSP and FitNet over Hinton, highlights the importance of better model initialization.
    \item Online distillation is a promising direction which removes the necessity of having a large pre-trained teacher for supervision and instead relies on mutual learning between a cohort of student models collectively supervising each other. This highlights the effectiveness of collaborative learning in improving the generalization of the models.
\end{enumerate}


\subsection{Label Noise}\label{label_noise}
Much of the success of the supervised learning methods can be attributed to the availability of huge amounts of high-quality annotations \cite{deng2009imagenet}.
However, real-world datasets often contain a certain amount of label noise arising from the difficulty of manual annotation. Furthermore, to leverage  the large amount of open-sourced data, automated label generation methods \cite{tsai2008automatically,mozafari2014scaling} are employed which make use of user tags and keywords which inherently leads to noisy labels.
The performance of the model is significantly affected by label noise \cite{sukhbaatar2014training,zhang2016understanding} and studies have shown that DNNs have the capacity to memorize noisy labels \cite{arpit2017closer}.
It is therefore pertinent for the training procedure to be more robust to label noise and efficiently learn under noisy supervision.

One reason for the failure of standard training is that the only supervision the model receives is the one-hot-labels. Therefore, when the ground-truth label is incorrect, the model doesn't receive any other useful information to learn from. 
In KD on the other hand, in addition to the ground truth label, the model receives supervision from the teacher, e.g. the soft probabilities in case of Hinton provides useful information about the relative probabilities amongst the classes. Similarly, in online distillation, the consensus between different students provides extra supervision.
We hypothesize that these extra supervision signals in the KD framework can mitigate the adverse effect of incorrect ground truth labels.

To test our hypothesis, we simulate label corruption on CIFAR-10 dataset whereby for each training image, we corrupt the true label with a given probability (referred to as noise rate, $\sigma$) to a randomly chosen class sampled from a uniform distribution on the number of classes\cite{hendrycks2019using}. We test the robustness of the various KD methods for different noise rates, $\sigma\in\{0.2,0.4,0.6\}$. We use WRN-16-2 as student and WRN-40-2 as teacher and follow the same training procedure (Table \ref{tab:train-params}).

Table \ref{tab:label_noise} shows that majority of the KD methods improve the generalization of the student trained under varying degrees of label corruption over the baseline. Hinton proves to be an effective method for learning under label noise and provides substantial gains in performance. For a lower noise level (0.2), it provides the highest generalization. As observed in section \ref{gen_perf}, FitNet and FSP show similar behavior to Hinton.
Amongst the online distillation methods, ONE improves the generalization over the baseline for lower noise levels but significantly harms the performance for higher noise level (0.6). DML is amongst the more effective methods and provides performance comparable to Hinton. The performance of these methods demonstrate the effectiveness of soft-targets in providing useful information about the data points with corrupted labels and mitigating the adverse effect of noisy labels.
Among the relational knowledge distillation methods, SP interestingly performs markedly better for lower noise levels, which suggests that the pairwise angular information can be useful in learning efficiently under label noise. The results show that the efficacy of KD extends beyond model compression and offers a general-purpose learning framework which is more robust to noisy labels prevalent in real-world datasets than standard training.

\begin{table}[tb]
\centering
\caption{\footnotesize Test set performance (\%) on CIFAR-10 with different label noise rates, $\sigma$. The best results are in bold, and the results below the baseline are colored in blue. We run each experiment for 5 different seeds and report the mean $\pm$ 1 STD.}
\label{tab:label_noise}
\resizebox{\columnwidth}{!}{
\begin{tabular}{|l|cccc|}
\hline
\textbf{$\sigma$} & \textbf{0}       & \textbf{0.2}     & \textbf{0.4}     & \textbf{0.6}     \\ \hline \hline
\textbf{Baseline} & 93.95$\pm$0.18   & 79.44$\pm$0.29   & 64.47$\pm$1.06   & 47.84$\pm$1.81   \\ \hline
\textbf{Hinton}   & 94.28$\pm$0.09   & \bf 87.23$\pm$0.26   & 76.32$\pm$0.87   & 58.18$\pm$0.35   \\
\textbf{BSS}      & 94.27$\pm$0.18   & 80.28$\pm$0.33   & 71.46$\pm$0.20   & \textcolor{blue}{47.69$\pm$0.37}   \\\hline
\textbf{FitNet}   & 94.34$\pm$0.11   & 87.01$\pm$0.27   & \bf 76.73$\pm$0.52   & 58.12$\pm$1.00   \\
\textbf{FSP}      & 94.31$\pm$0.08   & 87.14$\pm$0.38   & 76.47$\pm$0.24   & 58.07$\pm$0.55   \\
\textbf{AT}       & 94.50$\pm$0.18   & 79.59$\pm$0.47   & \textcolor{blue}{64.46$\pm$0.88}   & \textcolor{blue}{46.44$\pm$0.78}   \\ \hline
\textbf{SP}       & \bf 94.64$\pm$0.17   & 83.77$\pm$0.61   & 70.32$\pm$0.76   & 49.46$\pm$0.57   \\
\textbf{RKD-D}    & 94.42$\pm$0.15   & 79.94$\pm$0.59   & \textcolor{blue}{64.05$\pm$0.47}   & 48.37$\pm$1.62   \\
\textbf{RKD-A}    & 94.62$\pm$0.14   & 80.26$\pm$0.33   & 64.61$\pm$1.04   & 47.94$\pm$1.14   \\
\textbf{RKD-DA}   & 94.52$\pm$0.11   & 80.45$\pm$0.58   & 65.10$\pm$1.08   & 48.90$\pm$0.52   \\ \hline
\textbf{ONE}      & \textcolor{blue}{92.80$\pm$0.08}   & 83.76$\pm$0.40   & 68.64$\pm$0.53   & \textcolor{blue}{40.49$\pm$1.12}   \\
\textbf{DML}      & 94.38$\pm$0.15   & 85.63$\pm$0.33   & 76.33$\pm$0.32   & \bf 59.89$\pm$1.66   \\ \hline
\end{tabular}}
\end{table}

\subsection{Class Imbalance}\label{class_imbalance}
In addition to noisy labels, high class imbalance is naturally inherent in many real-world applications wherein, the dataset is not uniformly distributed and some classes are more abundant than others \cite{van2018inaturalist}.
Models train with standard training exhibit bias towards the prevalent classes at the expense of minority classes.
One drawback of the standard training is that the model receives no information about a particular class other than the data points belonging to it. The model does not receive any information about the similarities between data points of different classes which can be useful in learning better representation for the minority classes. 
KD framework, on the contrary, provides additional relational information between the different classes, e.g. the relative probabilities of each class provided as soft targets or the pairwise similarities between data points belonging to different classes. We hypothesize that this additional relational information can be useful in learning the minority classes better. 

To test our hypothesis, we simulate varying degrees of class imbalance on the CIFAR-10 dataset. We follow the analysis performed in \cite{hendrycks2019using} whereby similar to Dong et al.\cite{dong2018imbalanced}, we employ the power law model to simulate class imbalance. The number of training samples for a class $c$ as follows,
$n_c = \nint{a/(b+(c-1)^{-\gamma}}\qquad$, where $\nint{.}\qquad$ is the integer rounding function, $\gamma$ represents an imbalance ratio, $a$ and $b$ are offset parameters to specify the largest and smallest class sizes. The training data becomes a power law class distribution as the imbalance ratio $\gamma$ decreases. We test the performance of the KD methods on varying degrees of imbalance; $\gamma\in\{0.2, 0.6, 1.0, 2.0\}$ and ($a, b$) are set so that the maximum and minimum class counts are 5000 and 250 respectively.  

Table \ref{tab:class-imbalance} shows that apart from ONE and BSS ($\gamma=2$), all the other KD methods improve the generalization performance of the baseline model. In particular, RKD-A and RKD-DA provide the highest generalization across all the $\gamma$ values followed by Hinton and DML.
The empirical results show KD methods offer a more effective approach for training models on imbalanced datasets. We emphasize that the efficacy of KD goes much beyond a model compression technique and it should be considered as a general-purpose training paradigm which offers more robustness to common challenges in the real-world datasets compared to the standard training procedure.

\begin{table}[tbh]
\centering
\caption{\footnotesize Test set performance (\%) on CIFAR-10 with different class imbalance rates, $\gamma$. The best results are in bold, and the results below the baseline are colored in blue. We run each experiment for 5 different seeds and report the mean $\pm$ 1 STD.}
\label{tab:class-imbalance}
\resizebox{\columnwidth}{!}{
\begin{tabular}{|l|cccc|}
\hline
\bf $\gamma$      & \textbf{0.20}    & \textbf{0.60}    & \textbf{1}       & \textbf{2}       \\\hline\hline
\textbf{Baseline} & 78.05$\pm$0.58   & 78.83$\pm$0.41   & 80.09$\pm$0.38   & 83.33$\pm$0.24   \\\hline
\textbf{Hinton}   & 79.15$\pm$0.28   & 80.08$\pm$0.25   & 81.18$\pm$0.51   & 83.69$\pm$0.69   \\
\textbf{BSS}      & 78.07$\pm$0.20   & 79.22$\pm$0.53   & 80.44$\pm$0.24   & \textcolor{blue}{82.15$\pm$0.22}   \\\hline
\textbf{FitNet}   & 79.14$\pm$0.28   & 80.07$\pm$0.37   & 81.15$\pm$0.32   & 83.55$\pm$0.32   \\
\textbf{FSP}      & 79.26$\pm$0.43   & 80.03$\pm$0.50   & 81.12$\pm$0.43   & 83.60$\pm$0.25   \\
\textbf{AT}       & 79.13$\pm$0.40   & 80.51$\pm$0.23   & 80.96$\pm$0.18   & 84.13$\pm$0.32   \\\hline
\textbf{SP}       & 78.21$\pm$0.73   & 79.44$\pm$0.29   & 80.33$\pm$0.50   & \textcolor{blue}{83.08$\pm$0.29}   \\
\textbf{RKD-D}    & 79.12$\pm$0.26   & 80.57$\pm$0.45   & 81.48$\pm$0.57   & 84.13$\pm$0.42   \\
\textbf{RKD-A}    & \bf 79.52$\pm$0.51   & 80.54$\pm$0.17   & \bf 81.52$\pm$0.36   & \bf 84.33$\pm$0.42   \\
\textbf{RKD-DA}   & 79.43$\pm$0.41   & \bf 80.63$\pm$0.20   & 81.50$\pm$0.37   & 84.02$\pm$0.21   \\\hline
\textbf{ONE}      & \textcolor{blue}{77.48$\pm$1.05}   & \textcolor{blue}{78.04$\pm$0.86}   & \textcolor{blue}{79.48$\pm$0.39}   & \textcolor{blue}{80.88$\pm$1.05}   \\
\textbf{DML}      & 78.99$\pm$0.33   & 80.34$\pm$0.66   & 81.33$\pm$0.31   & 84.06$\pm$0.42 \\\hline
\end{tabular}}
\end{table}

\begin{figure*}[tbh]
\centering
\includegraphics[scale=0.7]{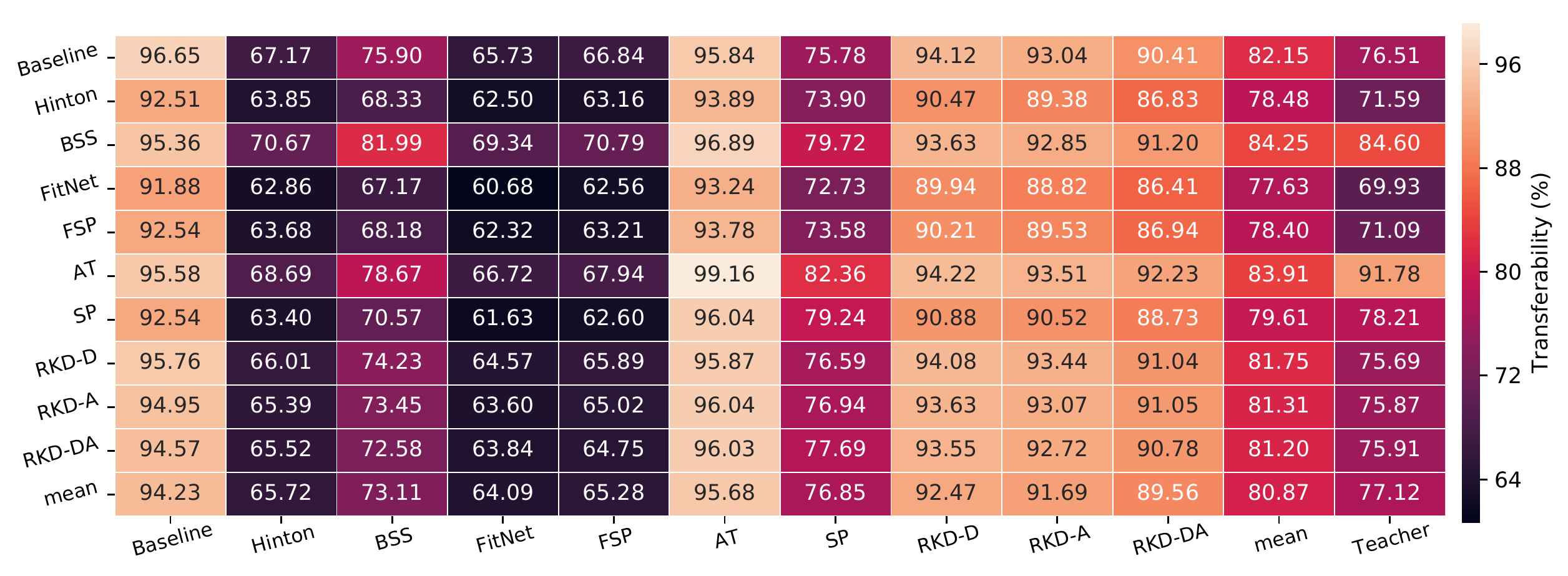}
\caption{Transferability of adversarial examples from source (columns) to target (rows) models.}
\label{fig:transferability}
\end{figure*}

\subsection{Transferability}\label{transferability}
To further study the characteristics of the different distillation methods and analyze how well they are able to mimic the teacher, we propose to use the transferability \cite{tramer2017space} of adversarial examples \cite{szegedy2013intriguing} as an approximation of the similarity of the decision boundary and representation space of the two models.
For each method, we first generate adversarial examples using the Projected Gradient Descent (PGD) method \cite{madry2017towards} with $\epsilon = 0.031$, $\eta = 0.003$ and $K = 20$. Then we perform a more fine-grained evaluation, by conducting the attack only on the subset of the test set which is correctly classified by both the source and target model and then reporting the success rate of the attack.
Higher transferability values indicate a higher level of similarity between the two models. 

Figure \ref{fig:transferability} (last column) shows the transferability of the adversarial examples generated on the teacher to students trained with different distillation methods.
AT achieves significantly higher transferability compared to other KD methods, suggesting that mimicking the latent representation space enables the student to be more similar to the teacher.
AT is followed by BSS which explicitly attempts to match the decision boundary of the teacher and therefore has much higher transferability compared to Hinton.
Relational Knowledge Distillation methods provide higher transferability than Hinton, FSP and FitNet even though they do not explicitly mimic the representation space or the decision boundary which suggest that maintaining the relations between data points implicitly causes the internal representation to be similar.
The comparatively lower values for Hinton and related methods (FitNet and FSP) might be explained by the effect of the Hinton method in improving the robustness of the model \cite{papernot2015distillation}. 

Figure \ref{fig:transferability} further shows the transferability of adversarial examples generated for each of the KD methoda to the other KD methods. The adversarial examples generated with AT, on average, provide the highest transferability to the other methods. This might be explained by its higher level of similarity to the teacher. Relational knowledge distillation methods achieve higher transferability whereas the methods which include the Hinton loss (response distillation methods, FitNet, and FSP) provide significantly lower transferability. On the other hand, these methods show more robustness to black-box attacks generated by the other methods.

\section{Conclusion}\label{conclusion}
In this study, we provided an extensive evaluation of nine different knowledge distillation methods and demonstrated the effectiveness and versatility of the KD framework. We studied the effect of mimicking different aspects of the teacher on the performance of the model and derived insights to guide the design of more effective KD methods.
We further showed the effectiveness of the KD framework in learning efficiently under varying degrees of label noise and class imbalance. Our study emphasizes that knowledge distillation should not only be considered as an efficient model compression technique but rather as a general-purpose training paradigm that offers more robustness to common challenges in the real-world datasets compared to the standard training procedure.

\bibliographystyle{IEEEtran}
\bibliography{References}

\begin{thebibliography}{10}
\providecommand{\url}[1]{#1}
\csname url@samestyle\endcsname
\providecommand{\newblock}{\relax}
\providecommand{\bibinfo}[2]{#2}
\providecommand{\BIBentrySTDinterwordspacing}{\spaceskip=0pt\relax}
\providecommand{\BIBentryALTinterwordstretchfactor}{4}
\providecommand{\BIBentryALTinterwordspacing}{\spaceskip=\fontdimen2\font plus
\BIBentryALTinterwordstretchfactor\fontdimen3\font minus
  \fontdimen4\font\relax}
\providecommand{\BIBforeignlanguage}[2]{{%
\expandafter\ifx\csname l@#1\endcsname\relax
\typeout{** WARNING: IEEEtran.bst: No hyphenation pattern has been}%
\typeout{** loaded for the language `#1'. Using the pattern for}%
\typeout{** the default language instead.}%
\else
\language=\csname l@#1\endcsname
\fi
#2}}
\providecommand{\BIBdecl}{\relax}
\BIBdecl

\bibitem{krizhevsky2012imagenet}
A.~Krizhevsky, I.~Sutskever, and G.~E. Hinton, ``Imagenet classification with
  deep convolutional neural networks,'' in \emph{Advances in neural information
  processing systems}, 2012, pp. 1097--1105.

\bibitem{voulodimos2018deep}
A.~Voulodimos, N.~Doulamis, A.~Doulamis, and E.~Protopapadakis, ``Deep learning
  for computer vision: A brief review,'' \emph{Computational intelligence and
  neuroscience}, vol. 2018, 2018.

\bibitem{zhou2017stochastic}
Z.~Zhou, P.~Mertikopoulos, N.~Bambos, S.~Boyd, and P.~W. Glynn, ``Stochastic
  mirror descent in variationally coherent optimization problems,'' in
  \emph{Advances in Neural Information Processing Systems}, 2017, pp.
  7040--7049.

\bibitem{han2015deep}
S.~Han, H.~Mao, and W.~J. Dally, ``Deep compression: Compressing deep neural
  networks with pruning, trained quantization and huffman coding,'' \emph{arXiv
  preprint arXiv:1510.00149}, 2015.

\bibitem{hinton2015distilling}
G.~Hinton, O.~Vinyals, and J.~Dean, ``Distilling the knowledge in a neural
  network,'' \emph{arXiv preprint arXiv:1503.02531}, 2015.

\bibitem{cheng2017survey}
Y.~Cheng, D.~Wang, P.~Zhou, and T.~Zhang, ``A survey of model compression and
  acceleration for deep neural networks,'' \emph{arXiv preprint
  arXiv:1710.09282}, 2017.

\bibitem{hu2019understanding}
W.~Hu, Z.~Li, and D.~Yu, ``Understanding generalization of deep neural networks
  trained with noisy labels,'' \emph{arXiv preprint arXiv:1905.11368}, 2019.

\bibitem{arpit2017closer}
D.~Arpit, S.~Jastrz{\k{e}}bski, N.~Ballas, D.~Krueger, E.~Bengio, M.~S. Kanwal,
  T.~Maharaj, A.~Fischer, A.~Courville, Y.~Bengio \emph{et~al.}, ``A closer
  look at memorization in deep networks,'' in \emph{Proceedings of the 34th
  International Conference on Machine Learning-Volume 70}.\hskip 1em plus 0.5em
  minus 0.4em\relax JMLR. org, 2017, pp. 233--242.

\bibitem{johnson2019survey}
J.~M. Johnson and T.~M. Khoshgoftaar, ``Survey on deep learning with class
  imbalance,'' \emph{Journal of Big Data}, vol.~6, no.~1, p.~27, 2019.

\bibitem{bucilua2006model}
C.~Buciluǎ, R.~Caruana, and A.~Niculescu-Mizil, ``Model compression,'' in
  \emph{Proceedings of the 12th ACM SIGKDD international conference on
  Knowledge discovery and data mining}.\hskip 1em plus 0.5em minus 0.4em\relax
  ACM, 2006, pp. 535--541.

\bibitem{heo2019knowledge}
B.~Heo, M.~Lee, S.~Yun, and J.~Y. Choi, ``Knowledge distillation with
  adversarial samples supporting decision boundary,'' in \emph{Proceedings of
  the AAAI Conference on Artificial Intelligence}, vol.~33, 2019, pp.
  3771--3778.

\bibitem{romero2014fitnets}
A.~Romero, N.~Ballas, S.~E. Kahou, A.~Chassang, C.~Gatta, and Y.~Bengio,
  ``Fitnets: Hints for thin deep nets,'' \emph{arXiv preprint arXiv:1412.6550},
  2014.

\bibitem{yim2017gift}
J.~Yim, D.~Joo, J.~Bae, and J.~Kim, ``A gift from knowledge distillation: Fast
  optimization, network minimization and transfer learning,'' in
  \emph{Proceedings of the IEEE Conference on Computer Vision and Pattern
  Recognition}, 2017, pp. 4133--4141.

\bibitem{zagoruyko2016paying}
S.~Zagoruyko and N.~Komodakis, ``Paying more attention to attention: Improving
  the performance of convolutional neural networks via attention transfer,''
  \emph{arXiv preprint arXiv:1612.03928}, 2016.

\bibitem{park2019relational}
W.~Park, D.~Kim, Y.~Lu, and M.~Cho, ``Relational knowledge distillation,'' in
  \emph{Proceedings of the IEEE Conference on Computer Vision and Pattern
  Recognition}, 2019, pp. 3967--3976.

\bibitem{tung2019similarity}
F.~Tung and G.~Mori, ``Similarity-preserving knowledge distillation,''
  \emph{arXiv preprint arXiv:1907.09682}, 2019.

\bibitem{zhang2018deep}
Y.~Zhang, T.~Xiang, T.~M. Hospedales, and H.~Lu, ``Deep mutual learning,'' in
  \emph{Proceedings of the IEEE Conference on Computer Vision and Pattern
  Recognition}, 2018, pp. 4320--4328.

\bibitem{lan2018knowledge}
X.~Lan, X.~Zhu, and S.~Gong, ``Knowledge distillation by on-the-fly native
  ensemble,'' in \emph{Proceedings of the 32nd International Conference on
  Neural Information Processing Systems}.\hskip 1em plus 0.5em minus
  0.4em\relax Curran Associates Inc., 2018, pp. 7528--7538.

\bibitem{krizhevsky2010cifar}
A.~Krizhevsky, V.~Nair, and G.~Hinton, ``Cifar-10 (canadian institute for
  advanced research),'' \emph{URL http://www. cs. toronto. edu/kriz/cifar.
  html}, vol.~8, 2010.

\bibitem{he2015deep}
K.~He, X.~Zhang, S.~Ren, and J.~Sun, ``Deep residual learning for image
  recognition. computer vision and pattern recognition (cvpr),'' in \emph{2016
  IEEE Conference on}, vol.~5, 2015, p.~6.

\bibitem{zagoruyko2016wide}
S.~Zagoruyko and N.~Komodakis, ``Wide residual networks,'' \emph{arXiv preprint
  arXiv:1605.07146}, 2016.

\bibitem{deng2009imagenet}
J.~Deng, W.~Dong, R.~Socher, L.-J. Li, K.~Li, and L.~Fei-Fei, ``Imagenet: A
  large-scale hierarchical image database,'' in \emph{2009 IEEE conference on
  computer vision and pattern recognition}.\hskip 1em plus 0.5em minus
  0.4em\relax Ieee, 2009, pp. 248--255.

\bibitem{tsai2008automatically}
C.-F. Tsai and C.~Hung, ``Automatically annotating images with keywords: A
  review of image annotation systems,'' \emph{Recent Patents on Computer
  Science}, vol.~1, no.~1, pp. 55--68, 2008.

\bibitem{mozafari2014scaling}
B.~Mozafari, P.~Sarkar, M.~Franklin, M.~Jordan, and S.~Madden, ``Scaling up
  crowd-sourcing to very large datasets: a case for active learning,''
  \emph{Proceedings of the VLDB Endowment}, vol.~8, no.~2, pp. 125--136, 2014.

\bibitem{sukhbaatar2014training}
S.~Sukhbaatar, J.~Bruna, M.~Paluri, L.~Bourdev, and R.~Fergus, ``Training
  convolutional networks with noisy labels,'' \emph{arXiv preprint
  arXiv:1406.2080}, 2014.

\bibitem{zhang2016understanding}
C.~Zhang, S.~Bengio, M.~Hardt, B.~Recht, and O.~Vinyals, ``Understanding deep
  learning requires rethinking generalization,'' \emph{arXiv preprint
  arXiv:1611.03530}, 2016.

\bibitem{van2018inaturalist}
G.~Van~Horn, O.~Mac~Aodha, Y.~Song, Y.~Cui, C.~Sun, A.~Shepard, H.~Adam,
  P.~Perona, and S.~Belongie, ``The inaturalist species classification and
  detection dataset,'' in \emph{Proceedings of the IEEE conference on computer
  vision and pattern recognition}, 2018, pp. 8769--8778.

\bibitem{hendrycks2019using}
D.~Hendrycks, K.~Lee, and M.~Mazeika, ``Using pre-training can improve model
  robustness and uncertainty,'' \emph{arXiv preprint arXiv:1901.09960}, 2019.

\bibitem{dong2018imbalanced}
Q.~Dong, S.~Gong, and X.~Zhu, ``Imbalanced deep learning by minority class
  incremental rectification,'' \emph{IEEE transactions on pattern analysis and
  machine intelligence}, vol.~41, no.~6, pp. 1367--1381, 2018.

\bibitem{tramer2017space}
F.~Tram{\`e}r, N.~Papernot, I.~Goodfellow, D.~Boneh, and P.~McDaniel, ``The
  space of transferable adversarial examples,'' \emph{arXiv preprint
  arXiv:1704.03453}, 2017.

\bibitem{szegedy2013intriguing}
C.~Szegedy, W.~Zaremba, I.~Sutskever, J.~Bruna, D.~Erhan, I.~Goodfellow, and
  R.~Fergus, ``Intriguing properties of neural networks,'' \emph{arXiv preprint
  arXiv:1312.6199}, 2013.

\bibitem{madry2017towards}
A.~Madry, A.~Makelov, L.~Schmidt, D.~Tsipras, and A.~Vladu, ``Towards deep
  learning models resistant to adversarial attacks,'' \emph{arXiv preprint
  arXiv:1706.06083}, 2017.

\bibitem{papernot2015distillation}
N.~Papernot, P.~D. McDaniel, X.~Wu, S.~Jha, and A.~Swami, ``Distillation as a
  defense to adversarial perturbations against deep neural networks. corr
  abs/1511.04508 (2015),'' in \emph{37th IEEE Symposium on Security and
  Privacy}, 2015.

\end{thebibliography}


\end{document}